\documentclass[10pt, a4paper]{article}

\usepackage[]{lrec-coling2024} 

\def\@fnsymbol#1{\ensuremath{\ifcase#1\or *\or \dagger\or \ddagger\or
   \mathsection\or \mathparagraph\or \|\or **\or \dagger\dagger
   \or \ddagger\ddagger \else\@ctrerr\fi}}
\newcommand{\ssymbol}[1]{^{\@fnsymbol{#1}}}

\title{Improving Legal Judgement Prediction in Romanian \\with Long Text Encoders}

\name{Mihai Masala$\ssymbol{1}$, Traian Rebedea$\ssymbol{1}$, Horia Velicu$\ssymbol{2}$} 

\address{$\ssymbol{1}$University Politehnica of Bucharest, $\ssymbol{2}$BRD Groupe Societe Generale \\
         mihai\_dan.masala@upb.ro, traian.rebedea@upb.ro, horia.velicu@brd.ro\\
}

\abstract{
In recent years,
the entire field of Natural Language Processing (NLP) has enjoyed amazing novel results 
achieving almost human-like performance on a variety of tasks. Legal NLP domain has also been part of this process, as it has seen an impressive growth. However, general-purpose models are not readily applicable for legal domain. Due to the nature of the domain (e.g. specialized vocabulary, long documents) specific models and methods are often needed for Legal NLP. In this work we investigate both specialized and general models for predicting the final ruling of a legal case, task known as Legal Judgment Prediction (LJP). We particularly focus on methods to extend to sequence length of Transformer-based models to better understand the long documents present in legal corpora. Extensive experiments on 4 LJP datasets in Romanian, originating from 2 sources with significantly different sizes and document lengths, show that specialized models and handling long texts are critical for a good performance. 
 \\ \newline \Keywords{legal judgement prediction, long context encoding, Romanian language} }

\begin{document}

\maketitleabstract

\section{Introduction}

The Transformer architecture~\cite{vaswani2017attention} initially proposed for machine translation has become almost ubiquitous for many Machine Learning tasks. Transformer based architectures~\cite{devlin2018bert,lewis2019bart} are used to develop state-of-the-art solution in a variety of fields, ranging from Natural Language Processing~\cite{sun2020mixup,devaraj2022evaluating} to Computer Vision~\cite{dosovitskiy2020image,patrick2021keeping}, Audio Signal Processing~\cite{radford2023robust} and image/video synthesis~\cite{ding2022cogview2,ge2022long}. Recently, Large Language Models~\cite{brown2020language} became capable of understanding and producing human-like text, leading to the advent of powerful conversational agents~\cite{touvron2023llama,vicuna2023,ouyang2022training}. Besides capable of engaging in human-like conversations, due to the huge amounts of pre-training and fine-tuning data, LLMs obtain state-of-the-art results on several tasks~\cite{openai2023gpt4}. 

Nevertheless, especially for highly specialized domains, there is still a need for custom models and methods. As such, legal~\cite{chalkidis2020legal,shao2020bert,masala2021jurbert,niklaus2022budgetlongformer,cui2023chatlaw}, medical~\cite{lee2020biobert,rasmy2021med,9628032}, chemical~\cite{chithrananda2020chemberta,ahmad2022chemberta} or financial~\cite{yang2020finbert,hillebrand2022kpi,wu2023bloomberggpt,yang2023fingpt} models have been proposed for a variety of languages. 

In this work we investigate how to effectively process the long documents in the legal domain for a low-resource language (Romanian). We experiment with four different datasets, provided by a one of the top banks in Romania, from two different sources. 
We are, to the best of our knowledge, the first to prove that SLED~\cite{ivgi2023efficient} encoding applied on long documents for the legal judgement prediction tasks significantly improves performance compared to baseline methods. This is especially important for low-resource languages, such as Romanian, where language-specific LLMs with long-context support are not yet available and existing multi-language LLMs have low performance, at least for Romanian as this study demonstrates.

The main takeaways from the experiments are that: 1) encoding long documents with SLED can provide an important increase of performance, 2) multi-lingual LLMs are currently under-performing on LJP in Romanian both on smaller and larger documents. 


\section{Related Work}

Transformer~\cite{vaswani2017attention} architectures use self-attention as a central component. This mechanism connects all tokens in a sequence in a graph-like manner, using a relatedness pooling operation. While powerful, self-attention comes at a great cost as it has a quadratic 
complexity with the input length. As documents in the legal domain can be very long, scaling the self-attention to such documents quickly becomes infeasible.

Therefore, a great amount of work has been done to address this limitation. One such category of solutions tries to reduce the quadratic complexity of the self-attention mechanism by restricting the number of tokens a particular token can attend to. In sparse attention, each token can and is influenced by a constant number of tokens, based on fixed~\cite{child2019generating,ainslie2020etc,zaheer2020big,beltagy2020longformer} or learned patterns~\cite{kitaev2020reformer,roy2021efficient}. Usually, a small constant number of global tokens (attending all the other tokens) are kept at each layer. 

Longformer~\cite{beltagy2020longformer} makes use of dilated sliding windows enabling long-range coverage while keeping sparsity. This is accomplished by having gaps in the attention patterns, increasing them as the model goes deeper. Accordingly, lower levels have strong local patterns while higher levels are capable of modeling long-range interactions. Finally, global attention is added on a small number of fixed input locations. 

Instead of trying to increase the effective sequence length, SLED~\cite{ivgi2023efficient} proposes an efficient way of splitting the text into smaller blocks with partial overlap to allow longer sequences to be encoded. This mechanism is akin to local attention, as "full" self-attention is applied in each block. We adapt this mechanism for classification tasks, generating a representation for each token, representations that are further aggregated and fed to a decision layer. 

Transformer-based models already assist legal practitioners on a multitude of tasks such as judgement prediction~\cite{chalkidis2019neural,huang2021dependency}, information extraction~\cite{chen2020joint,hendrycks2021cuad} or text classification~\cite{chalkidis2019large,chalkidis2020empirical}. Popular benchmarks devised for the legal domain~\cite{chalkidis2021lexglue,niklaus2023lextreme} usually contain long documents, beyond the maximum length of standard BERT-like models. Popular approaches~\cite{niklaus2022empirical} split a document into equal-length blocks and encode them separately. All the obtained embeddings are further fed into another Transformer, followed by a max-pooling operation, thus obtaining an embedding for the document. This method first builds context-unaware paragraph representations that are further contextualized at paragraph level by the second stage Transformer.

\section{Experimental Setup}
\subsection{Datasets}

\begin{table*}[h]
\centering
\begin{tabular}{|l|c|c|c|c|}
\hline
\textbf{Data source} & \textbf{Size} & \textbf{Class balance} & \textbf{jurBERT \#tokens} & \textbf{Llama2 \#tokens} \\
\hline
BankingCases ADM & 14367 & 1.59:1 & 2201 / 1161 & 3115 / 1684 \\
BankingCases ENF & 15044 & 1.51:1 & 2374 / 1225 & 3561 / 1874 \\
\hline
BRDCases ADM & 236 & 1.11:1 & 14280 / 10952 & 24047 / 17358 \\
BRDCases ENF & 90 & 3.10:1 & 6536 / 4270 & 10601 / 6912 \\
\hline

\end{tabular}
\caption{Dataset statistics - for number of tokens, the mean and median are shown for each tokenizer.}
\label{table:dataset_statistics}
\end{table*}

All datasets that we employ stem from Romanian civil cases in which clients sue a banking institution. Given the client's plea the task is to determine the outcome of the case.
We treat this task as a binary classification task (win for client or bank). In this work we use two sources that contain different types of documents for the cases. The first data source is a collection of banking cases that took place between 2010 and 2018. We will further refer to this corpus as \textbf{BankingCases}. Each case contains the summary of the plaintiff written by the judge presiding over the case. In most cases, the judge restructures and rewrites the original arguments, even distorting some arguments to make the ruling more convincing. While this adds a certain bias and does not represent a realistic use case, using such data as an intermediate finetuning dataset greatly increases performance on real-world scenarios~\cite{masala2021jurbert}. 

Finally, we collect a set of real-world cases, \textbf{BRDCases} provided by the juridical department of bank BRD Group Societe 
Generale. Compared to BankingCases, these documents represent the plaintiff's raw plea, a collection of requests, proofs and other relevant documents. We pass all documents through a specialized OCR in Romanian juridical domain and anonymize personal information. For this reasons, the dataset contains less structured data, longer documents and more noise stemming, in part, due to the nature of the OCR extraction process.

From both sources we extract two common types of cases of interest to the banking domain, namely administration fee litigations (ADM) and enforcement appeals (ENF). In Table~\ref{table:dataset_statistics} we present detailed statistics for each dataset employed in this work. Note the large discrepancy between  both the number of samples and the length of each case between BankingCases and BRDCases. In the real-world setting (BRDCases) we have extremely long texts, very few samples and in the case of ENF rather unbalanced data. For all cases, we automatically extract the year and the county where the case was filed. We inject this information in the form of one-hot encoding after the Transformer output, before the final decision and we further refer to it as handcrafted features. 

To summarize, we collect datasets from two sources. The first dataset (BankingCases) contains a set of cases where the input is represented by the summary of arguments of both sides provided by the judge presiding the case at the end of the trial. The second dataset (BRDCases) contains a set of real-world argumentation of the plaintiff submitted to the court at the beginning of the trial, in exactly the same format they are received by the legal department of the bank. This means that for BRDCases, the input contains only the arguments of one side (i.e. the plaintiff), consists of much longer documents that come in the form of scanned files that need to be digitized. Our main goal is to provide an efficient automated method for predicting the outcome of a case in this real-world scenario. Such a method allows legal teams to efficiently assign resources, filtering out \textit{unwinnable} cases.
\subsection{Models and Approaches}

We employ a variety of methods to adapt the jurBERT model~\cite{masala2021jurbert}
to handle texts longer than 512 tokens. The first and simplest method is to make jurBERT process more than one block of 512 tokens in parallel. Therefore we experiment with the first and last 512 tokens of a document (denoted as \textit{2*512}); similar for the first, middle, and last 512 tokens (denoted as \textit{3*512}). Aggregating results from multiple blocks of a document is done by concatenating the [CLS] token representation for each block. This approach allows for handling longer documents, does not add a lot of complexity, and keeps the running time low. However, it is a rather rudimentary approach as it treats different blocks completely independently as there is no self-attention between blocks.

Next, we build Longformer versions of jurBERT, increasing the maximum sequence length and adapting the attention mechanism. This effectively increases the maximum sequence length of the model, and we experiment with sizes up to 4096. We also adapt SLED~\cite{ivgi2023efficient} input preprocessing for our task (just dropping the decoder part), feeding the model a total of 32 chunks of 256 tokens (denoted as \textit{32*256}).

Recent Large Language Models are already pretrained using large contexts. Llama2~\cite{touvron2023llama} is multi-lingual LLM with a context length of 4096, while the Romanian Okapi~\cite{lai2023okapi}, a version of Llama~\cite{touvron2023llama1}, shares the same maximum sequence length.

Previous work has shown that Transformer-based solutions outperform several other approaches such as LSTMs, CNNs or SVMs with String Kernels~\cite{lodhi2002text} for Romanian legal judgement prediction in a very similar setting~\cite{masala2021jurbert}. Compared to the datasets used by \citet{masala2021jurbert}, we have collected a larger set of real-world cases and we pre-process them with several tools for the Romanian language that have shown an improvement in accuracy (i.e. a specialized Romanian juridical domain OCR extractor, a personal identifiable information anonymizer and a Romanian diacritics restoration tool). Overall, our real-world data is greater in size, more diverse and less noisy. For these reasons, in this work we decide to limit our experiments to the best performing Transformer-based models from ~\cite{masala2021jurbert} as a baseline and to show a significant improvement over them with long-context support.

\subsection{Training setup}

Each model is trained using 5-fold cross-validation, over a maximum of 10 epochs. After each epoch, we save the AUC on the current "test" split and select the final result as the highest mean (over all folds) AUC for each epoch. Due to computational limits, for BankingCases we take only one run, while for BRDCases we run each model 3 times (for a total of 15 runs). 

Note that all experiments on BRDCases are using models that are first finetuned on BankingCases sharing the same model architecture and hyperparameters. For computational reasons we finetune Llama2 and Okapi models on BankingCases using only a sequence length of 1024. 

\section{Results and Discussions}

\begin{table}[h]
\centering
\begin{tabular}{|l|c|r|r|}
\hline
\textbf{Model} & \textbf{Seq Len} &  \textbf{Mean AUC} & \textbf{Std AUC} \\
\hline
jurBERT & 512 & 78.20 & 0.56\\
jurBERT & 2*512  & \textbf{78.37} & 1.05\\
jurBERT & 3*512& 78.50 & 1.16\\
jurBERT$\ssymbol{2}$& 1024 & 74.27 & 0.73\\

jurBERT$\ssymbol{2}$& 2048 & 70.65 & 3.02\\
jurBERT$\ssymbol{2}$& 4096& 67.20 & 0.92\\
\hline
jurBERT & 32*256$\ssymbol{3}$& 67.57 & 0.66\\
\hline
jurBERT$\ssymbol{1}$& 512 & 78.13 & 1.02\\
Llama2$\ssymbol{1}$& 1024  & 69.88 & 0.79 \\
Okapi$\ssymbol{1}$& 1024 & 69.66 & 1.20 \\
\hline
\end{tabular}
\caption{Results on BankingCases ADM. $\ssymbol{1}$ marks models that do not use handcrafted features, $\ssymbol{2}$ marks Longformer variants and $\ssymbol{3}$ marks SLED input. We mark the top performer with \textbf{bold}.}
\label{table:results_adm_big}
\end{table}

\begin{table}[h]
\centering
\begin{tabular}{|l|c|r|r|}
\hline
\textbf{Model} & \textbf{Seq Len} & \textbf{Mean AUC} & \textbf{Std AUC} \\
\hline
jurBERT & 512 & 75.26 & 0.56\\
jurBERT & 2*512 & \textbf{78.57} & 0.42\\
jurBERT & 3*512 & 77.93 & 0.59\\
jurBERT$\ssymbol{2}$&1024& 66.76 & 3.36\\

jurBERT$\ssymbol{2}$&2048& 56.33 & 5.26\\
jurBERT$\ssymbol{2}$&4096& 54.24 & 2.56\\
\hline
jurBERT & 32*256$\ssymbol{3}$& 78.03 & 0.76\\
\hline
jurBERT$\ssymbol{1}$& 512 & 75.08 & 0.47\\
Llama2$\ssymbol{1}$& 1024 & 65.03 & 2.19 \\
Okapi$\ssymbol{1}$& 1024 & 64.11 & 2.16 \\
\hline
\end{tabular}
\caption{Results on BankingCases ENF. Notations are the same as in Table~\ref{table:results_adm_big}.}
\label{table:results_exec_big}
\end{table}

\begin{table}[h]
\centering
\begin{tabular}{|l|c|r|r|}
\hline
\textbf{Model} & \textbf{Seq Len} & \textbf{Mean AUC} & \textbf{Std AUC} \\
\hline
jurBERT & 512 & 68.38 & 5.49\\
jurBERT & 2*512 & 64.28 & 4.55\\
jurBERT & 3*512 & 63.15 & 7.42\\
jurBERT$\ssymbol{2}$& 1024 & 71.33 & 7.38\\
jurBERT$\ssymbol{2}$& 2048 & 71.55 & 5.25\\
jurBERT$\ssymbol{2}$& 4096 & 71.56 & 5.38\\
\hline
jurBERT & 32*256$\ssymbol{3}$& \textbf{72.71} & 5.99\\
\hline
jurBERT$\ssymbol{1}$& 512 & 62.73 & 4.82\\
Llama2$\ssymbol{1}$& 1024 & 63.60 & 6.93 \\
Okapi$\ssymbol{1}$& 1024 & 61.35 & 6.74 \\
\hline
\end{tabular}
\caption{Results on BRDCases ADM. Notations are the same as in Table~\ref{table:results_adm_big}.}
\label{table:results_adm_small}
\end{table}

\begin{table}[h]
\centering
\begin{tabular}{|l|c|r|r|}
\hline
\textbf{Model} & \textbf{Seq Len} & \textbf{Mean AUC} & \textbf{Std AUC} \\
\hline
jurBERT & 512 & 63.80 & 11.25\\
jurBERT & 2*512 & \textbf{69.63} & 10.37\\
jurBERT & 3*512 & 60.54 & 11.32\\
jurBERT$\ssymbol{2}$& 1024 & 60.87 & 10.14\\
jurBERT$\ssymbol{2}$& 2048 & 56.92 & 13.81\\
jurBERT$\ssymbol{2}$& 4096 & 41.60 & 22.79\\
\hline
jurBERT &32*256$\ssymbol{3}$& 65.48 & 12.01\\
\hline
jurBERT$\ssymbol{1}$& 512 & 60.53 & 8.16\\
Llama2$\ssymbol{1}$& 1024 & 60.19 & 12.42 \\
Okapi$\ssymbol{1}$& 1024 & 63.56 & 11.22 \\
\hline
\end{tabular}
\caption{Results on BRDCases ENF. Notations are the same as in Table~\ref{table:results_adm_big}.}
\label{table:results_exec_small}
\end{table}

The results on BankingCases are presented in Table~\ref{table:results_adm_big} and Table~\ref{table:results_exec_big}. The top part of the tables contain results for vanilla and Longformer variants of jurBERT. In the middle section of the tables the SLED alternative is introduced, while in the last section results using LLMs are showed. Note that results in the bottom part of tables do not use handcrafted features. 

The first thing to note is the strong performance of the jurBERT baseline with a maximum sequence length of 512. For the Longformer variants, we believe their lack of performance is due to the limited training data (15k total, 12k training samples) that does not allow the model to properly learn how to handle longer sequences. In the case of BankingCases, as documents are basically summaries written by judges, in most cases the strongest argument in favor of the final ruling is present in the first part of the document. This is in stark contrast with arguments of the plaintiff where the order and even the quality of documents is not always "best first". Understanding, validating, and ranking such arguments requires highly specialized work that is done by the judge.

The rather limited training data problem is aggravated in the case of LLMs. Both Llama2 and Okapi are both general language models, not specialized in the legal domain. This is also clearly visible by the statistics about the number of tokens presented in Table~\ref{table:dataset_statistics}. jurBERT uses a specialized vocabulary (in Romanian juridical domain) and therefore is much more efficient in encoding legal texts compared to the general multi-language vocabulary used by Llama2/Okapi models. Furthermore, both Llama2 and Okapi have been trained on very few texts in Romanian.

Interestingly, in a setting with extremely low number of documents that are also very long (BRDCases dataset), the hierarchy of models changes. As seen in Table~\ref{table:results_adm_small}, processing longer sequences generates better results, with the SLED variant obtaining the best result. 
For enforcement appeals (Table~\ref{table:results_exec_small}), we find jurBERT with first and last 512 tokens yields the best performance. Also, due to very limited and unbalanced data (only 90 samples, with a 3.10:1 distribution) note the very high standard deviation values. In some extreme cases, the model is unable to learn on some folds, leading to extremely poor results (under 0.5 mean AUC). As for each fold only 72 samples are used for training and the evaluation is performed only on 18 samples, the standard deviation is very high for most models.

\section{Conclusions}

In this work we investigated the applicability of language models on the task of Legal Judgement Prediction, in a low-resource language (i.e. Romanian). We proved that integrating longer sequences, especially using SLED-style encoding, allows for a better understanding of documents, leading in the end to a overall increase in performance in our low-resources and long-documents setting. Experiments on four different datasets highlight the need for methods that allow language models to parse long sequences and specialized vocabularies. As seen in Table~\ref{table:dataset_statistics}, a specialized vocabulary is more efficient in encoding such documents, effectively allowing more information to be processed under the same sequence length limit. But a long sequence length is not enough. Especially in the case of BRDCases, the more relevant dataset of the two as it represents a real-world scenario, a long context size does not guarantee a competitive performance with Longformer and Llama2/Okapi variants underperforming and SLED offering the only improvement. At the same time, LLMs trained on huge amounts of (multi-lingual) data still lag behind more specialized solutions in this low-resource setting. 

\section{Limitations and Ethical Statement}
In this work, we employ legal judgement prediction mainly to help one of the sides in a trial, in this case the defendant (a bank) to understand its chances of winning a trial. We do not aim to substitute the juridical process and, at the same time, understand that having such a system might provide important additional information for the side using it. 

Our work focuses on a low-resource language and uses (very) small datasets. Moreover, the BRDCases dataset might have some biases as it contains legal documents received from a single Romanian bank. Therefore, the results presented in the paper might not be relevant for other languages, might not transfer to different tasks or even data from other parties on the same task. 

On the other hand, this scenario is very relevant and useful for the legal department of a large bank, and we consider that this scenario is of interested for other researchers working on real-world datasets and use-cases.

While legal documents contain personal identifiable (PII), we want to highlight that in our experiments PII data has been removed using an external API for Romanian. Again, we consider that this pre-processing is important to remove any spurious correlations and might also be relevant for other real-world use-cases.

\nocite{*}
\section{Bibliographical References}\label{sec:reference}

\bibliographystyle{lrec-coling2024-natbib}
\bibliography{lrec-coling2024-example}

\end{document}